\documentclass[sigconf]{acmart}

\usepackage{booktabs} 

\usepackage{algorithm}
\usepackage{algpseudocode}

\setcopyright{rightsretained}

\acmDOI{none}

\acmISBN{none}

\acmConference[KDD'18]{ACM}{July 2018}{London, UK}

\renewcommand\footnotetextcopyrightpermission[1]{} 
\begin{document}

\title{Consistent Individualized Feature Attribution for Tree Ensembles}

\author{Scott M. Lundberg, Gabriel G. Erion,  and Su-In Lee}
\affiliation{%
  \institution{University of Washington}
}
\email{{slund1,erion,suinlee}@uw.edu}

\begin{abstract}
Interpreting predictions from tree ensemble methods such as gradient boosting machines and random forests is important, yet feature attribution for trees is often heuristic and not individualized for each prediction. Here we show that popular feature attribution methods are \emph{inconsistent}, meaning they can lower a feature's assigned importance when the true impact of that feature actually increases. This is a fundamental problem that casts doubt on any comparison between features. To address it we turn to recent applications of game theory and develop fast exact tree solutions for SHAP (\underline{SH}apley \underline{A}dditive ex\underline{P}lanation) values, which are the unique consistent and locally accurate attribution values.  We then extend SHAP values to interaction effects and define {\it SHAP interaction values}. We propose a rich visualization of individualized feature attributions that improves over classic attribution summaries and partial dependence plots, and a unique ``supervised'' clustering (clustering based on feature attributions). We demonstrate better agreement with human intuition through a user study, exponential improvements in run time, improved clustering performance, and better identification of influential features. An implementation of our algorithm has also been merged into XGBoost and LightGBM, see \url{http://github.com/slundberg/shap} for details.

\end{abstract}

\maketitle


\section{Introduction}

Understanding why a model made a prediction is important for trust, actionability, accountability, debugging, and many other tasks. To understand predictions from tree ensemble methods, such as gradient boosting machines or random forests, importance values are typically attributed to each input feature. These importance values can be computed either for a single prediction (individualized), or an entire dataset to explain a model's overall behavior (global).

\begin{figure*}
  \centering
  \includegraphics[width=0.99\textwidth]{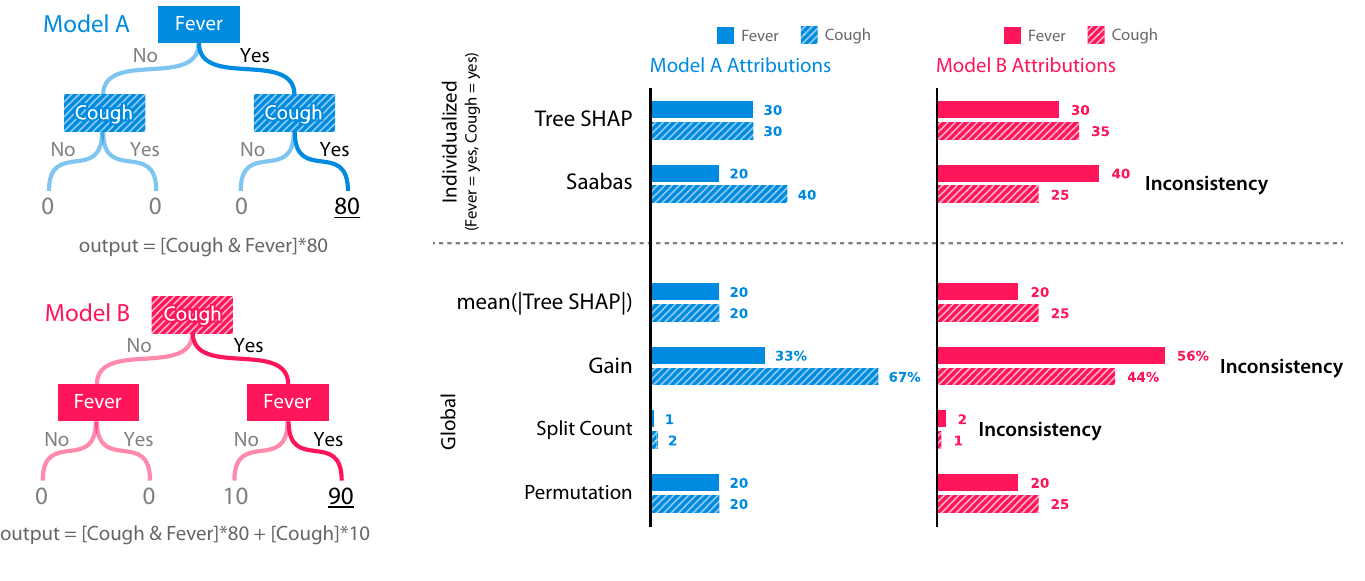}
  \caption[Simple example]{Two simple tree models that demonstrate inconsistencies in the Saabas, gain, and split count attribution methods: The Cough feature has a larger impact in Model B than Model A, but is attributed less importance in Model B. Similarly, the Cough feature has a larger impact than Fever in Model B, yet is attributed less importance. The individualized attributions explain a single prediction of the model (when both Cough and Fever are Yes) by allocating the difference between the expected value of the model's output (20 for Model A, 25 for Model B) and the current output (80 for Model A, 90 for Model B). The global attributions represent the overall importance of a feature in the model. Without consistency it is impossible to reliably compare feature attribution values.
  }
  \label{fig:and_trees}
\end{figure*}

Concerningly, popular current feature attribution methods for tree ensembles are {\it inconsistent}. This means that when a model is changed such that a feature has a higher impact on the model's output, current methods can actually lower the importance of that feature. Inconsistency strikes at the heart of what it means to be a good attribution method, because it prevents the meaningful comparison of attribution values across features. 
This is because inconsistency implies that a feature with a large attribution value might be less important than another feature with a smaller attribution (see Figure \ref{fig:and_trees} and Section \ref{sec:current}).




To address this problem we turn to the recently proposed SHAP (SHapley Additive exPlanation) values \cite{lundberg2017unified}, which are based on a unification of ideas from game theory \cite{vstrumbelj2014explaining} and local explanations \cite{ribeiro2016should}. Here we show that by connecting tree ensemble feature attribution methods with the class of {\it additive feature attribution methods} \cite{lundberg2017unified} we can motivate SHAP values as the only possible consistent feature attribution method with several desirable properties.

SHAP values are theoretically optimal, but like other model agnostic feature attribution methods \cite{baehrens2010explain,ribeiro2016should,vstrumbelj2014explaining,datta2016algorithmic}, they can be challenging to compute. To solve this we derive an algorithm for tree ensembles that reduces the complexity of computing exact SHAP values from $O(T L2^M)$ to $O(T LD^2)$ where $T$ is the number of trees, $L$ is the maximum number of leaves in any tree, $M$ is the number of features, and $D$ is the maximum depth of any tree. This exponential reduction in complexity allows predictions from previously intractable models with thousands of trees and features to now be explained in a fraction of a second. Entire datasets can now be explained, which enables new alternatives to traditional partial dependence plots and feature importance plots \cite{friedman2001elements}, which we term {\it SHAP dependence plots} and {\it SHAP summary plots}, respectively. 

Current attribution methods cannot directly represent interactions, but must divide the impact of an interaction among each feature. To directly capture pairwise interaction effects we propose {\it SHAP interaction values}; an extension of SHAP values based on the Shapley interaction index from game theory \cite{fujimoto2006axiomatic}. SHAP interaction values bring the benefits of guaranteed consistency to explanations of interaction effects for individual predictions.

In what follows we first discuss current tree feature attribution methods and their inconsistencies. We then introduce SHAP values as the only possible consistent and locally accurate attributions, present Tree SHAP as a high speed algorithm for estimating SHAP values of tree ensembles, then extend this to SHAP interaction values. We use user study data, computational performance, influential feature identification, and supervised clustering to compare with previous methods. Finally, we illustrate SHAP dependence plots and SHAP summary plots with XGBoost and NHANES I national health study data \cite{miller1973plan}.

\section{Inconsistencies in current feature attribution methods}
\label{sec:current}

Tree ensemble implementations in popular packages such as XGBoost \cite{chen2016xgboost}, scikit-learn \cite{pedregosa2011scikit}, and the {\it gbm} R package \cite{ridgeway2010generalized} allow a user to compute a measure of feature importance. These values are meant to summarize a complicated ensemble model and provide insight into what features drive the model's prediction.

{\it Global} feature importance values are calculated for an entire dataset (i.e., for all samples) in three primary ways:
\begin{enumerate}
\item Gain: A classic approach to feature importance introduced by Breiman et al. in 1984 \cite{breiman1984classification} is based on gain. Gain is the total reduction of loss or impurity contributed by all splits for a given feature. Though its motivation is largely heuristic \cite{friedman2001elements}, gain is widely used as the basis for feature selection methods \cite{sandri2008bias, irrthum2010inferring, chebrolu2005feature}.
\item Split Count: A second common approach 
is simply to count how many times a feature is used to split \cite{chen2016xgboost}. Since feature splits are chosen to be the most informative, this can represent a feature's importance.
\item Permutation: A third common approach 
is to randomly permute the values of a feature in the test set and then observe the change in the model's error. If a feature's value is important then permuting it should create a large increase in the model's error. Different choices about the method of feature value permutation lead to variations of this basic approach \cite{strobl2008conditional,ishwaran2007variable, auret2011empirical,diaz2006gene,rodenburg2008framework}.
\end{enumerate}

{\it Individualized} methods that compute feature importance values for a single prediction are less established for trees. While model agnostic individualized explanation methods \cite{lundberg2017unified,baehrens2010explain,ribeiro2016should,vstrumbelj2014explaining,datta2016algorithmic} can be applied to trees \cite{lundberg2017explainable}, they are significantly slower than tree-specific methods and have sampling variability (see Section \ref{sec:experiments} for a computational comparison, or \cite{lundberg2017unified} for an overview). The only current tree-specific individualized explanation method we are aware of is by Sabbas \cite{path_blog}. The Saabas method is similar to the classic dataset-level gain method, but instead of measuring the reduction of loss, it measures the change in the model's expected output. It proceeds by comparing the expected value of the model output at the root of the tree with the expected output of the sub-tree rooted at the child node followed by the decision path of the current input. The difference between these expectations is then attributed to the feature split on at the root node. By repeating this process recursively the method allocates the difference between the expected model output and the current output among the features on the decision path.

\begin{figure*}
  \centering
  \includegraphics[width=0.9\textwidth]{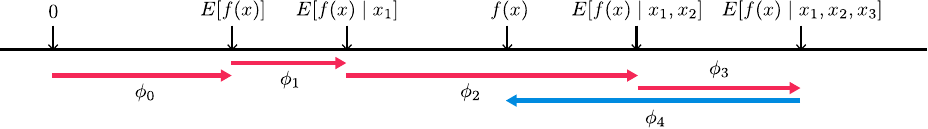}
  \caption{SHAP (\underline{SH}apley \underline{A}dditive ex\underline{P}lanation) values explain the output of a function $f$ as a sum of the effects $\phi_i$ of each feature being introduced into a conditional expectation. Importantly, for non-linear functions the order in which features are introduced matters. SHAP values result from averaging over all possible orderings. Proofs from game theory show this is the only possible consistent approach where $\sum_{i=0}^M \phi_i = f(x)$. In contrast, the only current individualized feature attribution method for trees satisfies the summation, but is inconsistent because it only considers a single ordering \cite{path_blog}.}
  \label{fig:number_line}
\end{figure*}

\emph{Unfortunately, the feature importance values from the gain, split count, and Saabas methods are all inconsistent.} This means that a model can change such that it relies more on a given feature, yet the importance estimate assigned to that feature decreases. Of the methods we consider, 
only SHAP values and permutation-based methods are consistent.
Figure \ref{fig:and_trees} shows the result of applying all these methods to two simple regression trees.\footnote{For clarity we rounded small values in Figure \ref{fig:and_trees}. These small values are why the lower left splits in both models were not pruned during training.} For the global calculations we assume an equal number of dataset points fall in each leaf, and the label of those points is exactly equal to the prediction of the leaf. Model~A represents a simple AND function, while Model~B represents the same AND function but with an additional increase in the predicted value when Cough is ``Yes''. Note that because Cough is now more important it gets split on first in Model~B.

Individualized feature attribution is represented by Tree~SHAP and Sabbas for the input Fever=Yes and Cough=Yes. Both methods allocate the difference between the current model output and the expected model output among the input features ($80 - 20$ for Model~A). But the SHAP values are guaranteed to reflect the importance of the feature (see Section~\ref{sec:shap}), while the Saabas values can give erroneous results, such as a larger attribution to Fever than to Cough in Model~B.

Global feature attribution is represented by four methods: the mean magnitude of the SHAP values, gain, split count, and feature permutation. Only the mean SHAP value magnitude and permutation correctly give Cough more importance than Fever in Model~B. This means gain and split count are not reliable measures of global feature importance, which is important to note given their widespread use.




\subsection{SHAP values as the only consistent and locally accurate individualized feature attributions}
\label{sec:shap}

It was recently noted that many current methods for interpreting individual machine learning model predictions fall into the class of {\it additive feature attribution methods} \cite{lundberg2017unified}. This class covers methods that explain a model's output as a sum of real values attributed to each input feature.

\begin{definition}
\label{def:additive}
{\bf Additive feature attribution methods} have an explanation model $g$ that is a linear function of binary variables:
\begin{equation}
\label{eq:additive_fa}
g(z') = \phi_0 + \sum_{i = 1}^M \phi_i z_i',
\end{equation}
where $z' \in \{0,1\}^M$, $M$ is the number of input features, and $\phi_i \in \mathbb{R}$.
\end{definition}

The $z'_i$ variables typically represent a feature being observed ($z'_i = 1$) or unknown ($z'_i = 0$), and the $\phi_i$'s are the feature attribution values.

As previously described in \citeauthor{lundberg2017unified} (2017), an important property of the class of additive feature attribution methods is that there is a single unique solution in this class with three desirable properties: {\it local accuracy}, {\it missingness}, and {\it consistency}. Local accuracy states that the sum of the feature attributions is equal to the output of the function we are seeking to explain. Missingness states that features that are already missing (such that $z'_i = 0$) are attributed no importance. Consistency states that changing a model so a feature has a larger impact on the model will never decrease the attribution assigned to that feature.



Note that in order to evaluate the effect missing features have on a model $f$, it is necessary to define a mapping $h_x$ that maps between a binary pattern of missing features represented by $z'$ and the original function input space. Given such a mapping we can evaluate $f(h_x(z'))$ and so calculate the effect of observing or not observing a feature (by setting $z'_i = 1$ or $z'_i = 0$).

To compute SHAP values we define $f_x(S) = f(h_x(z')) = E[f(x) \mid x_S]$ where $S$ is the set of non-zero indexes in $z'$ (Figure \ref{fig:number_line}), and $E[f(x) \mid x_S]$ is the expected value of the function conditioned on a subset $S$ of the input features. SHAP values combine these conditional expectations with the classic Shapley values from game theory to attribute $\phi_i$ values to each feature:

\begin{equation}
\label{eq:shapley}
\phi_i = \sum_{S \subseteq N \setminus \{i\}} \frac{|S|!(M - |S| -1)!}{M!} \left [ f_x(S \cup \{i\}) - f_x(S) \right ],
\end{equation}

\noindent where $N$ is the set of all input features.

As shown in \citeauthor{lundberg2017unified} (2017), the above method is the only possible consistent, locally accurate method that obeys the missingness property and uses conditional dependence to measure missingness \cite{lundberg2017unified}. This is strong motivation to use SHAP values for tree ensemble feature attribution, particularly since the only previous individualized feature attribution method for trees, the Saabas method, satisfies both local accuracy and missingness using conditional dependence, but fails to satisfy consistency. This means that SHAP values provide a strict theoretical improvement by eliminating significant consistency problems (Figure~\ref{fig:and_trees}).

\section{Tree SHAP: Fast SHAP value computation for trees}
\label{sec:tree_shap}

Despite the compelling theoretical advantages of SHAP values, their practical use is hindered by two problems:

\begin{enumerate}
\item The challenge of estimating $E[f(x) \mid x_S]$ efficiently.
\item The exponential complexity of Equation \ref{eq:shapley}.
\end{enumerate}

Here we focus on tree models and propose fast SHAP value estimation methods specific to trees and ensembles of trees. We start by defining a slow but straightforward algorithm, then present the much faster and more complex Tree~SHAP algorithm.

\subsection{Estimating SHAP values directly in $O(TL2^M)$ time}
\label{sec:shap_direct}

If we ignore computational complexity then we can compute the SHAP values for a tree by estimating $E[f(x) \mid x_S]$ and then using Equation \ref{eq:shapley} where $f_x(S) = E[f(x) \mid x_S]$. For a tree model $E[f(x) \mid x_S]$ can be estimated recursively using Algorithm \ref{alg:exp_value}, where $v$ is a vector of node values, which takes the value $internal$ for internal nodes. The vectors $a$ and $b$ represent the left and right node indexes for each internal node. The vector $t$ contains the thresholds for each internal node, and $d$ is a vector of indexes of the features used for splitting in internal nodes. The vector $r$ represents the cover of each node (i.e., how many data samples fall in that sub-tree). The weight $w$ measures what proportion of the training samples matching the conditioning set $S$ fall into each leaf.

\begin{algorithm}
\caption{Estimating $E[f(x) \mid x_S]$ \label{alg:exp_value}}
\begin{algorithmic}
\Procedure{EXPVALUE}{$x$, $S$, $tree = \{v, a, b, t, r, d\}$}
\Procedure{G}{$j$, $w$}
  \If{$v_j \ne internal$}
    \State \Return{$w \cdot v_j$}
  \Else
    \If{$d_j \in S$}
      \State \Return \Call{G}{$a_j$, $w$} {\bf if} $x_{d_j} \le t_j$ {\bf else} \Call{G}{$b_j$, $w$}
    \Else
    \State \Return \Call{G}{$a_j$, $w r_{a_j} / r_j$} + \Call{G}{$b_j$, $w r_{b_j} / r_j$}
    \EndIf
    
  \EndIf
\EndProcedure
\State \Return \Call{G}{$1$, $1$}
\EndProcedure
\end{algorithmic}
\end{algorithm}

\subsection{Estimating SHAP values in $O(TLD^2)$ time}
\label{sec:shap_fast}

Here we propose a novel algorithm to calculate the same values as above, but in polynomial time instead of exponential time. Specifically, we propose an algorithm that runs in $O(TLD^2)$ time and $O(D^2 + M)$ memory, where for balanced trees the depth becomes $D = \log L$. Recall $T$ is the number of trees, $L$ is the maximum number of leaves in any tree, and $M$ is the number of features.


The intuition of the polynomial time algorithm is to recursively keep track of what proportion of all possible subsets flow down into each of the leaves of the tree. 
This is similar to running Algorithm \ref{alg:exp_value} simultaneously for all $2^M$ subsets $S$ in Equation \ref{eq:shapley}. It may seem reasonable to simply keep track of how many subsets (weighted by the cover splitting of Algorithm \ref{alg:exp_value}) pass down each branch of the tree. However, this combines subsets of different sizes and so prevents the proper weighting of these subsets, since the weights in Equation \ref{eq:shapley} depend on $|S|$. To address this we keep track of each possible subset size during the recursion. The {\it EXTEND} method in Algorithm \ref{alg:tree_shap} grows all these subsets according to a given fraction of ones and zeros, while the {\it UNWIND} method reverses this process and is commutative with {\it EXTEND}. The {\it EXTEND} method is used as we descend the tree. The {\it UNWIND} method is used to undo previous extensions when we split on the same feature twice, and to undo each extension of the path inside a leaf to compute weights for each feature in the path.

\begin{algorithm}
\caption{Tree SHAP \label{alg:tree_shap}}
\begin{algorithmic}
\Procedure{TS}{$x$, $tree = \{v, a, b, t, r, d\}$}
\State $\phi = \textrm{array of $len(x)$ zeros}$
\Procedure{RECURSE}{$j$, $m$, $p_z$, $p_o$, $p_i$}
  \State $m =~$\Call{EXTEND}{$m$, $p_z$, $p_o$, $p_i$}
  \If{$v_j \ne internal$}
    \For{$i \gets 2 \textrm{ to } len(m)$}
      \State $w = sum(\Call{UNWIND}{m, i}.w)$
      \State $\phi_{m_i} = \phi_{m_i} + w(m_i.o - m_i.z) v_j$
    \EndFor
  \Else
    \State $h,c = x_{d_j} \le t_j~?~(a_j,b_j) : (b_j,a_j)$
    \State $i_z = i_o = 1$
    \State $k = \Call{FINDFIRST}{m.d, d_j}$
    \If{$k \ne \textrm{nothing}$}
      \State $i_z,i_o = (m_k.z,m_k.o)$
      \State $m = \Call{UNWIND}{m, k}$
    \EndIf
    \State \Call{RECURSE}{$h$, $m$, $i_z r_h/r_j$, $i_o$, $d_j$}
    \State \Call{RECURSE}{$c$, $m$, $i_z r_c/r_j$, $0$, $d_j$}
  \EndIf
\EndProcedure
\Procedure{EXTEND}{$m$, $p_z$, $p_o$, $p_i$}
\State $l = len(m)$
\State $m = copy(m)$
\State $m_{l+1}.(d,z,o,w) = (p_i, p_z, p_o, l = 0~?~1:0)$
\For{$i \gets l-1 \textrm{ to } 1$}
  \State $m_{i+1}.w = m_{i+1}.w + p_o m_i.w(i/l)$
  \State $m_i.w =  p_z m_i.w[(l-i)/l]$
\EndFor
\State \Return m
\EndProcedure
\Procedure{UNWIND}{$m$, $i$}
\State $l = len(m)$
\State $n = m_l.w$
\State $m = copy(m_{1...l-1})$
\For{$j \gets l-1 \textrm{ to } 1$}
  \If{$m_i.o \ne 0$}
    \State $t = m_j.w$
    \State $m_j.w = n \cdot l/(j \cdot m_i.o)$
    \State $n = t - m_j.w \cdot m_i.z ((l-j)/l)$
  \Else
    \State $m_j.w = (m_j.w \cdot l)/(m_i.z (l-j))$
  \EndIf
\EndFor
\For{$j \gets i \textrm{ to } l-1$}
  \State $m_j.(d,z,o) = m_{j+1}.(d,z,o)$
\EndFor
\State \Return m
\EndProcedure
\State \Call{RECURSE}{$1$, $[]$, $1$, $1$, $0$}
\State \Return $\phi$
\EndProcedure
\end{algorithmic}
\end{algorithm}

In Algorithm \ref{alg:tree_shap}, $m$ is the path of unique features we have split on so far, and contains four attributes: $d$ the feature index, $z$ the fraction of ``zero'' paths (where this feature is not in the set $S$) that flow through this branch, $o$ the fraction of ``one'' paths (where this feature is in the set $S$) that flow through this branch, and $w$ which is used to hold the proportion of sets of a given cardinality that are present. We use the dot notation to access these members, and for the whole vector $m.d$ represents a vector of all the feature indexes.

Algorithm \ref{alg:tree_shap} reduces the computational complexity of exact SHAP value computation from exponential to low order polynomial for trees and sums of trees (since the SHAP values of a sum of two functions is the sum of the original functions' SHAP values).





\section{SHAP Interaction Values}

Feature attributions are typically allocated among the input features, one for each feature, but we can gain additional insight by separating \emph{interaction effects} from main effects. If we consider pairwise interactions this leads to a matrix of attribution values representing the impact of all pairs of features on a given model prediction. Since SHAP values are based on classic Shapley values from game theory, a natural extension to interaction effects can be obtained though the more modern Shapley interaction index \cite{fujimoto2006axiomatic}:

\begin{equation}
\label{eq:shapley_interactions}
\Phi_{i,j} = \sum_{S \subseteq N \setminus \{i,j\}} \frac{|S|!(M - |S| - 2)!}{2(M - 1)!} \nabla_{ij}(S),
\end{equation}

\noindent when $i \ne j$, and
\begin{align}
\label{eq:shapley_interaction_grad}
\nabla_{ij}(S) &= f_x(S \cup \{i,j\}) - f_x(S \cup \{i\}) - f_x(S \cup \{j\}) + f_x(S) \\
&= f_x(S \cup \{i,j\}) - f_x(S \cup \{j\}) - [f_x(S \cup \{i\}) - f_x(S)]. \label{eq:shapley_interaction_grad2}
\end{align}

\noindent In Equation~\ref{eq:shapley_interactions} the SHAP interaction value between feature $i$ and feature $j$ is split equally between each feature so $\Phi_{i,j} = \Phi_{j,i}$ and the total interaction effect is $\Phi_{i,j} + \Phi_{j,i}$. The main effects for a prediction can then be defined as the difference between the SHAP value and the SHAP interaction values for a feature:

\begin{align}
\label{eq:shapley_interaction_main}
\Phi_{i,i} &= \phi_i - \sum_{j \ne i} \Phi_{i,j}.
\end{align}

These SHAP interaction values follow from similar axioms as SHAP values, and allow the separate consideration of main and interaction effects for individual model predictions. This separation can uncover important interactions captured by tree ensembles that might otherwise be missed (Figure \ref{fig:nhanes_sbp_main_and_age_interaction} in Section~\ref{sec:shap_plots}).

While SHAP interaction values can be computed directly from Equation~\ref{eq:shapley_interactions}, we can leverage Algorithm~\ref{alg:tree_shap} to drastically reduce their computational cost for tree models. As highlighted in Equation~\ref{eq:shapley_interaction_grad2} SHAP interaction values can be interpreted as the difference between the SHAP values for feature $i$ when feature $j$ is present and the SHAP values for feature $i$ when feature $j$ is absent. This allows us to use Algorithm~\ref{alg:tree_shap} twice, once while ignoring feature $j$ as fixed to present, and once with feature $j$ absent. This leads to a run time of $O(TMLD^2)$, since we repeat the process for each feature. Note that even though this computational approach does not seem to directly enforce symmetry, the resulting $\Phi$ matrix is always symmetric.

\section{Experiments and Applications}

We compare Tree SHAP and SHAP interaction values with previous methods through both traditional metrics and three new applications we propose for individualized feature attributions: supervised clustering, SHAP summary plots, and SHAP dependence plots.\footnote{Jupyter notebooks to compute all results are available at \url{http://github.com/slundberg/shap/notebooks/tree_shap_paper}}

\begin{figure}
  \centering
  \includegraphics[width=0.38\textwidth]{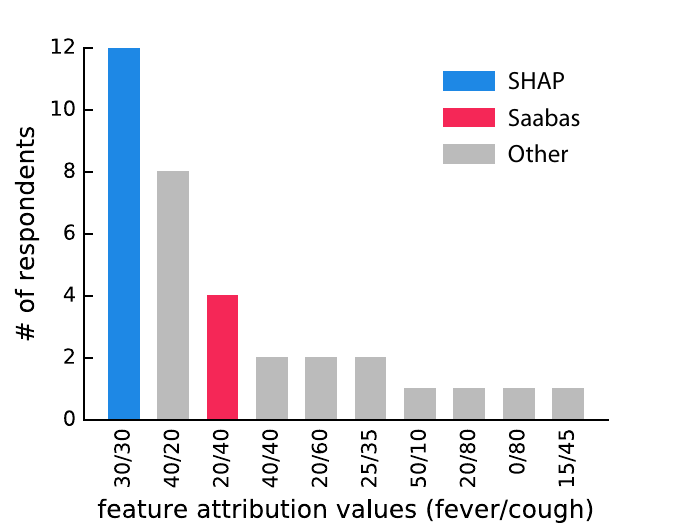}
  \caption{Feature attribution values from 34 participants shown the tree from Model A in Figure \ref{fig:and_trees}. The first number represents the allocation to the Fever feature, while the second represents the allocation to the Cough feature. Participants from Amazon Mechanical Turk were not selected for machine learning expertise. No constraints were placed on the feature attribution values users entered.}
  \label{fig:tree_max}
\end{figure}

\begin{figure*}
  \centering
  \includegraphics[width=1.0\textwidth]{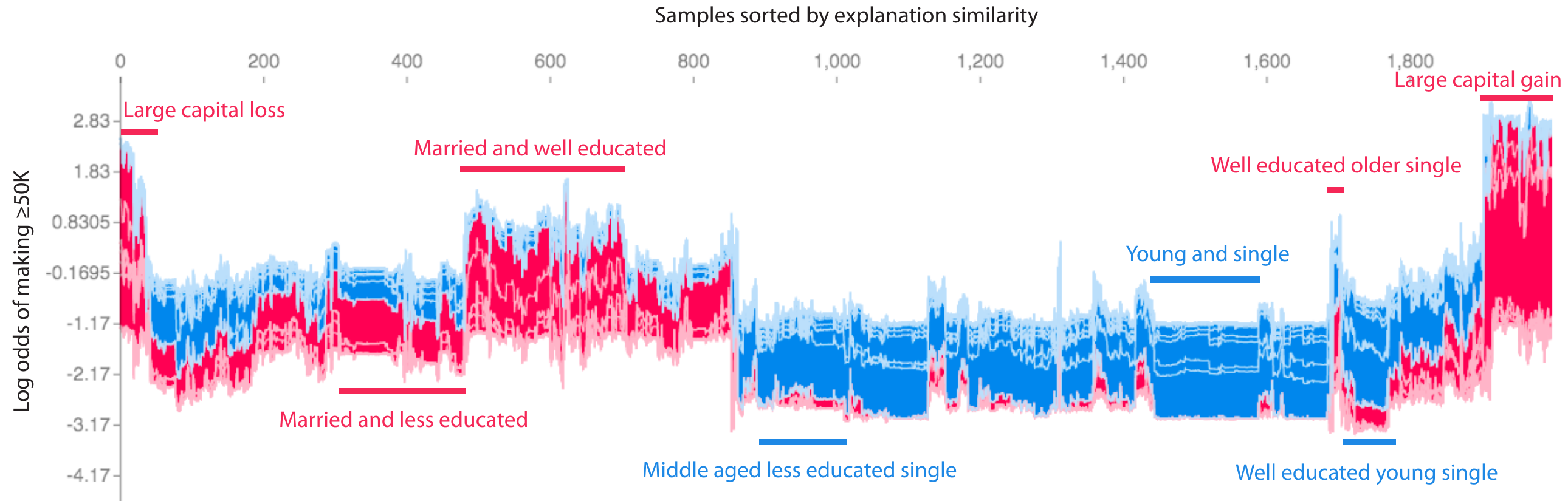}
  \caption{Supervised clustering with SHAP feature attributions in the UCI census dataset identifies among 2,000 individuals distinct subgroups of people that share similar reasons for making money. An XGBoost model with 500 trees of max depth six was trained on demographic data using a shrinkage factor of $\eta = 0.005$. This model was then used to predict the log odds that each person makes $\ge\$50K$. Each prediction was explained using Tree SHAP, and then clustered using hierarchical agglomerative clustering (imagine a dendrogram above the plot joining the samples). Red feature attributions push the score higher, while blue feature attributions push the score lower (as in Figure \ref{fig:number_line} but rotated 90$^\circ$). A few of the noticeable subgroups are annotated with the features that define them.}
  \label{fig:census_explanation}
\end{figure*}

\subsection{Agreement with Human Intuition}
\label{sec:user_study}

To validate that the SHAP values in Model A of Figure \ref{fig:and_trees} are the most natural assignment of credit we ran a user study to measure people's intuitive feature attribution values. Model A's tree was shown to participants and said to represent risk for a certain disease. They were told that when a given person was found to have both a cough and fever their risk went up from the prior risk of 20 (the expected value of risk) to a risk of 80. Participants were then asked to apportion the 60 point change in risk among the Cough and Fever features as they saw best.

Figure \ref{fig:tree_max} presents the results of the user study for Model A. The equal distribution of credit used by SHAP values was found to be the most intuitive. A smaller number of participants preferred to give greater weight to the first feature to be split on (Fever), while still fewer followed the allocation of the Saabas method and gave greater weight to the second feature split on (Cough).

\subsection{Computational Performance}
\label{sec:run_time}

\begin{figure}
  \centering
  \includegraphics[width=0.42\textwidth]{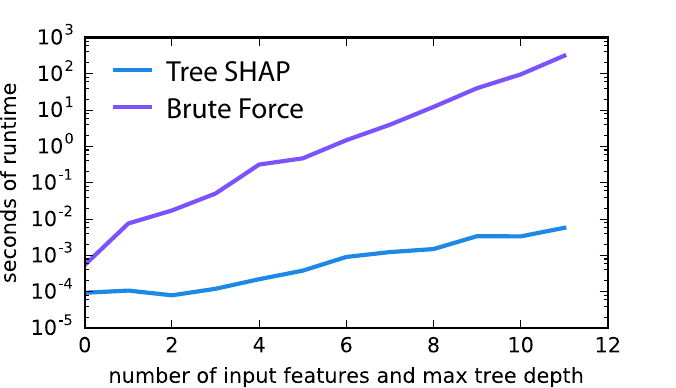}
  \caption{Runtime improvement of Algorithm 2 over using Equation \ref{eq:shapley} and Algorithm 1. An XGBoost model with 50 trees was trained using an equally increasing number of input features and max tree depths. The time to explain one input vector is reported.}
  \label{fig:runtime}
\end{figure}

Figure \ref{fig:runtime} demonstrates the significant run time improvement provided by Algorithm 2. Problems that were previously intractable for exact computation are now inexpensive. An XGBoost model with 1,000 depth 10 trees over 100 input features can now be explained in 0.08 seconds.

\subsection{Supervised Clustering}
\label{sec:experiments}
One intriguing application enabled by individualized feature attributions is what we term ``supervised clustering,'' where instead of using an unsupervised clustering method directly on the data features, you run clustering on the feature attributions.

Supervised clustering naturally handles one of the most challenging problems in unsupervised clustering: determining feature weightings (or equivalently, determining a distance metric). Many times we want to cluster data using features with very different units. 
Features may be in dollars, meters, unit-less scores, etc. but whenever we use them as dimensions in a single multidimensional space it forces any distance metric to compare the relative importance of a change in different units (such as dollars vs. meters). Even if all our inputs are in the same units, often some features are more important than others. Supervised clustering uses feature attributions to naturally convert all the input features into values with the same units as the model output. This means that a unit change in any of the feature attributions is comparable to a unit change in any other feature attribution. It also means that fluctuations in the feature values only effect the clustering if those fluctuations have an impact on the outcome of interest.

Here we demonstrate the use of supervised clustering on the classic UCI census dataset \cite{Lichman2013}. For this dataset the goal is to predict from basic demographic data if a person is likely to make more than \$50K annually.
By representing the positive feature attributions as red bars and the negative feature attributions as blue bars (as in Figure \ref{fig:number_line}), we can stack them against each other to visually represent the model output as their sum. Figure \ref{fig:census_explanation} does this vertically for predictions from 2,000 people from the census dataset. The explanations for each person are stacked horizontally according the leaf order of a hierarchical clustering of the SHAP values. This groups people with similar reasons for a predicted outcome together. The formation of distinct subgroups of people demonstrates the power of supervised clustering to identify groups that share common factors related to income level.

\begin{figure}[t!h]
  \centering
  \includegraphics[width=0.4\textwidth]{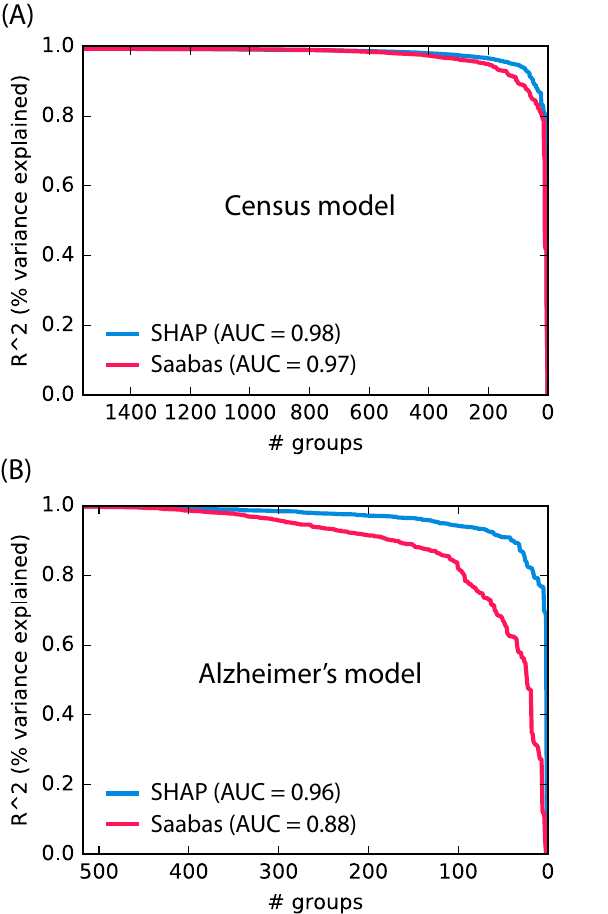}
  \caption{A quantitative measure of supervised clustering performance. If all samples are placed in their own group, and each group predicts the mean value of the group, then the $R^2$ value (the proportion of model output variance explained) will be $1$. If groups are then merged one-by-one the $R^2$ will decline until when there is only a single group it will be $0$. Hierarchical clusterings that well separate the model output value will retain a high $R^2$ longer during the merging process. Here supervised clustering with SHAP values outperformed the Sabbas method in both (A) the census data clustering shown in Figure \ref{fig:census_explanation}, and (B) a clustering from gene-based predictions of Alzheimer's cognitive scores.}
  \label{fig:cluster_plots}
\end{figure}

One way to quantify the improvement provided by SHAP values over the heuristic Saabas attributions is by examining how well supervised clustering based on each method explains the variance of the model output (note global feature attributions are not considered since they do not enable this type of supervised clustering). If feature attribution values well-represent the model then supervised clustering groups will have similar function outputs. Since hierarchical clusterings encode many possible groupings, we plot in Figure \ref{fig:cluster_plots} the change in the $R^2$ value as the number of groups shrinks from one group per sample ($R^2 = 1$) to a single group ($R^2 = 0$). For the census dataset, groupings based on SHAP values outperform those from Saabas values (Figure \ref{fig:cluster_plots}A). For a dataset based on cognitive scores for Alzheimer's disease SHAP values significantly outperform Saabas values (Figure \ref{fig:cluster_plots}B). This second dataset contains 200 gene expression module levels \cite{celik2014efficient} as features and CERAD cognitive scores as labels \cite{mirra1991consortium}.


\subsection{Identification of Influential Features}

\begin{figure}
  \centering
  \includegraphics[width=0.45\textwidth]{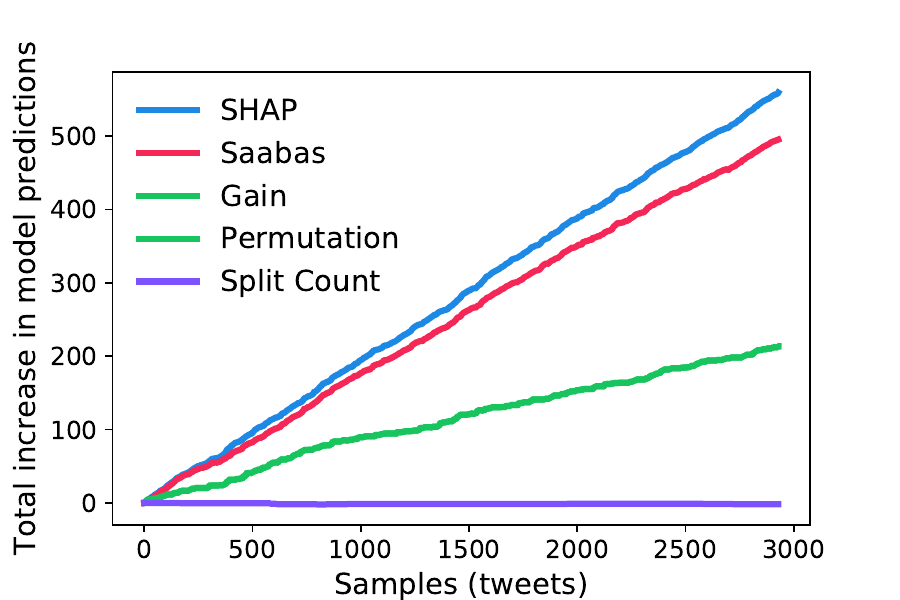}
  \caption{The total increase in a sentiment model's output when the most negative feature is replaced. Five different attribution methods were used to determine the most negative feature for each sample. The higher the total increase in model output, the more accurate the attribution method was at identifying the most influential negative feature.}
  \label{fig:sentiment_perturbation}
\end{figure}

Feature attribution values are commonly used to identify which features influenced a model's prediction the most. To compare methods, the change in a model's prediction can be computed when the most influential feature is perturbed. Figure~\ref{fig:sentiment_perturbation} shows the result of this experiment on a sentiment analysis model of airline tweets \cite{airline_tweets}. An XGBoost model with 50 trees of maximum depth 30 was trained on 11,712 tweets with 1,686 bag-of-words features. Each tweet had a sentiment score label between -1 (negative) and 1 (positive). The predictions of the XGBoost model were then explained for 2,928 test tweets. For each method we choose the most influential negative feature and replaced it with the value of the same feature in another random tweet from the training set (this is designed to mimic the feature being unknown). The new input is then re-run through the model to produce an updated output. If the chosen feature significantly lowered the model output, then the updated model output should be higher than the original. By tracking the total change in model output as we progress through the test tweets we observe that SHAP values best identify the most influential negative feature. Since global methods only select a single feature for the whole dataset we only replaced this feature when it would likely increase the sentiment score (for gain and permutation this meant randomly replacing the ``thank'' feature when it was missing, for split count it was the word ``to''). 


\subsection{SHAP Plots}
\label{sec:shap_plots}

Plotting the impact of features in a tree ensemble model is typically done with a bar chart to represent global feature importance, or a partial dependence plot to represent the effect of changing a single feature \cite{friedman2001elements}. However, since SHAP values are individualized feature attributions, unique to every prediction, they enable new, richer visual representations. 
{\it SHAP summary plots} replace typical bar charts of global feature importance, and {\it SHAP dependence plots} provide an alternative to partial dependence plots that better capture interaction effects.

To explore these visualizations we trained an XGBoost Cox proportional hazards model on survival data from the classic NHANES I dataset \cite{miller1973plan} using the NHANES I Epidemiologic Followup Study \cite{cox1997plan}. After selection for the presence of basic blood test data we obtained data for 9,932 individuals followed for up to 20 years after baseline data collection for mortality. Based on a 80/20 train/test split we chose to use 7,000 trees of maximum depth 3, $\eta = 0.001$, and $50\%$ instance sub-sampling. We then used these parameters and trained on all individuals to generate the final model.

\subsubsection{SHAP Summary Plots}

\begin{figure}
  \centering
  \includegraphics[width=0.48\textwidth]{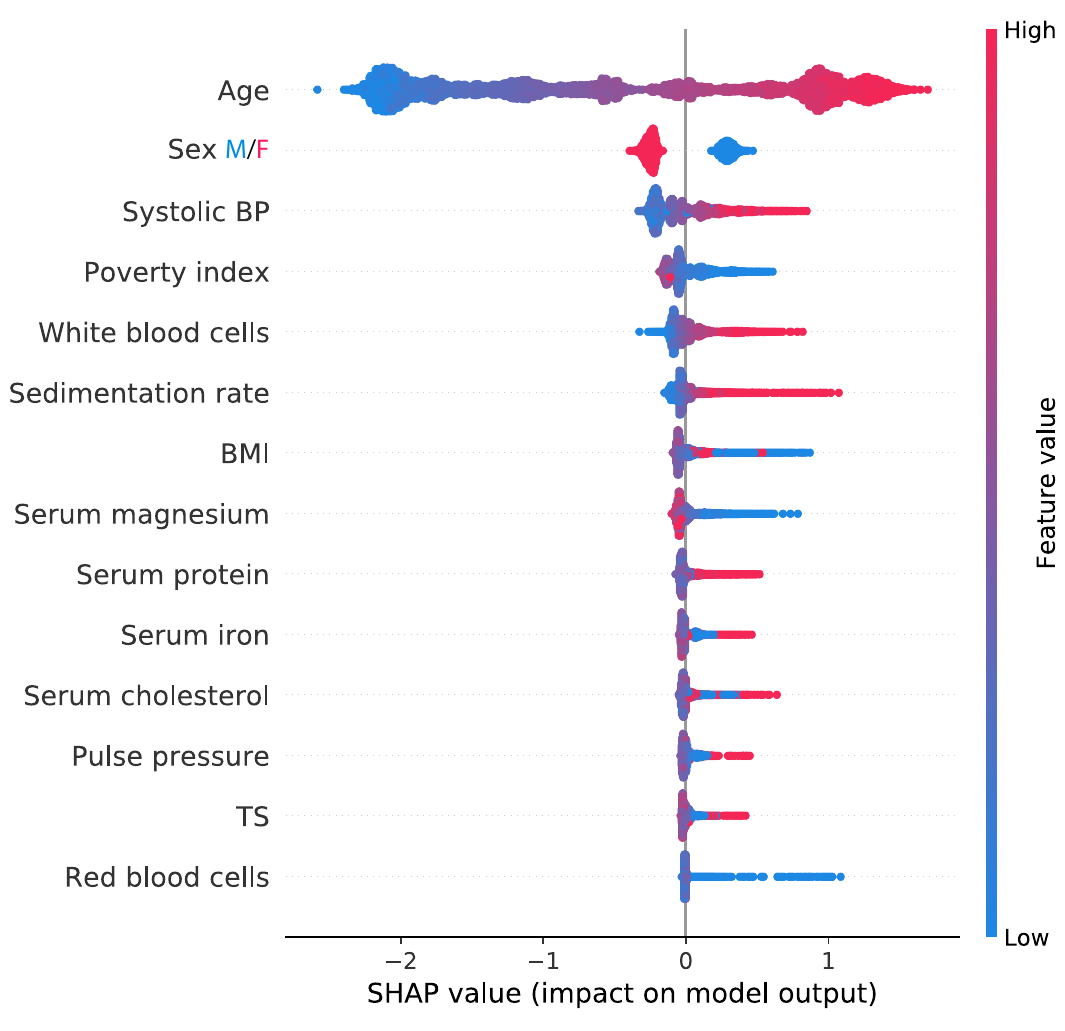}
  \caption{SHAP summary plot of a 14 feature XGBoost survival model on 20 year mortality followup data from NHANES I \cite{miller1973plan}. The higher the SHAP value of a feature, the higher your log odds of death in this Cox hazards model. Every individual in the dataset is run through the model and a dot is created for each feature attribution value, so one person gets one dot on each feature's line. Dot's are colored by the feature's value for that person and pile up vertically to show density.}
  \label{fig:nhanes_summary}
\end{figure}

Standard feature importance bar charts give a notion of relative importance in the training dataset, but they do not represent the range and distribution of impacts that feature has on the model's output, and how the feature's value relates to it's impact. SHAP summary plots leverage individualized feature attributions to convey all these aspects of a feature's importance while remaining visually concise (Figure~\ref{fig:nhanes_summary}). Features are first sorted by their global impact $\sum_{j=1}^N|\phi_i^{(j)}|$, then dots representing the SHAP values $\phi_i^{(j)}$ are plotted horizontally, stacking vertically when they run out of space. This vertical stacking creates an effect similar to violin plots but without an arbitrary smoothing kernel width. Each dot is colored by the value of that feature, from low (blue) to high (red). If the impact of the feature on the model's output varies smoothly as its value changes then this coloring will also have a smooth gradation. In Figure~\ref{fig:nhanes_summary} we see (unsurprisingly) that age at baseline is the most important risk factor for death over the next 20 years. The density of the age plot shows how common different ages are in the dataset, and the coloring shows a smooth increase in the model's output (a log odds ratio) as age increases. In contrast to age, systolic blood pressure only has a large impact for a minority of people with high blood pressure. The general trend of long tails reaching to the right, but not to the left, means that extreme values of these measurements can significantly raise your risk of death, but cannot significantly lower your risk.

\subsubsection{SHAP Dependence Plots}

As described in Equation 10.47 of \citeauthor{friedman2001elements} (2001), partial dependence plots represent the expected output of a model when the value of a specific variable (or group of variables) is fixed. The values of the fixed variables are varied and the resulting expected model output is plotted. Plotting how the expected output of a function changes as we change a feature helps explain how the model depends on that feature.

\begin{figure}
  \centering
  \includegraphics[width=0.45\textwidth]{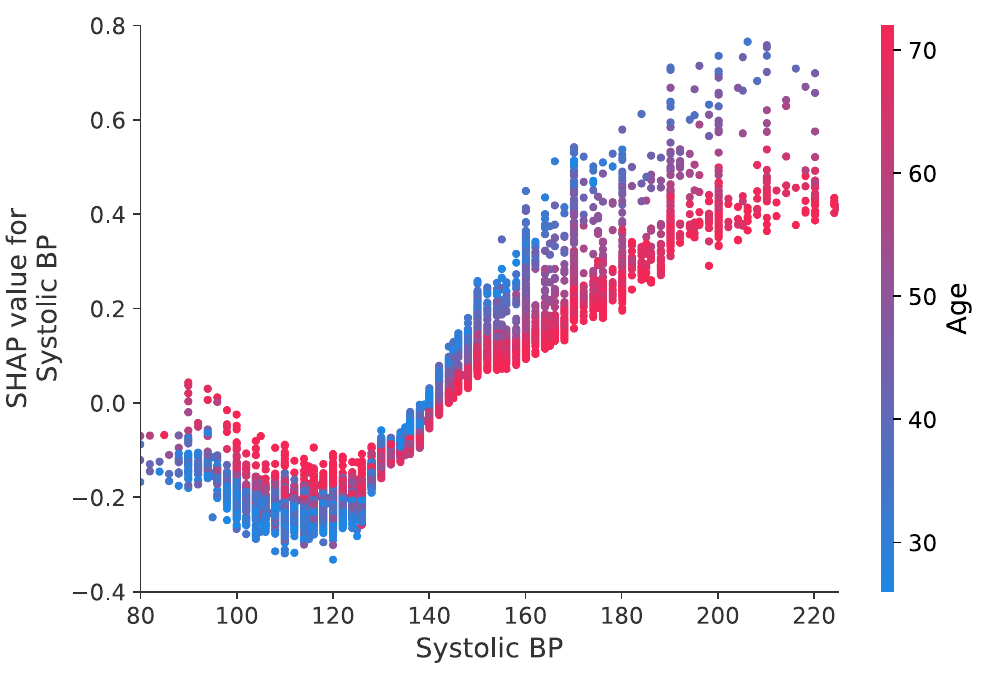}
  \caption{Each dot is a person. The x-axis is their systolic blood pressure and the y-axis is the SHAP value attributed to their systolic blood pressure. Higher SHAP values represent higher risk of death due to systolic blood pressure. Coloring each dot by the person's age reveals that high blood pressure is more concerning to the model when you are young (this represents an interaction effect).}
  \label{fig:nhanes_sbp}
\end{figure}

SHAP values can be used to create a rich alternative to partial dependence plots, which we term SHAP dependence plots. SHAP dependence plots use the SHAP value of a feature for the y-axis and the value of the feature for the x-axis. By plotting these values for many individuals from the dataset we can see how the feature's attributed importance changes as its value varies (Figure~\ref{fig:nhanes_sbp}). While standard partial dependence plots only produce lines, SHAP dependence plots capture vertical dispersion due to interaction effects in the model. These effects can be visualized by coloring each dot with the value of an interacting feature. In Figure~\ref{fig:nhanes_sbp} coloring by age shows that high blood pressure is more alarming when you are young. Presumably because it is both less surprising as you age, and possibly because it takes time for high blood pressure to lead to fatal complications.

\begin{figure}
  \centering
  \includegraphics[width=0.45\textwidth]{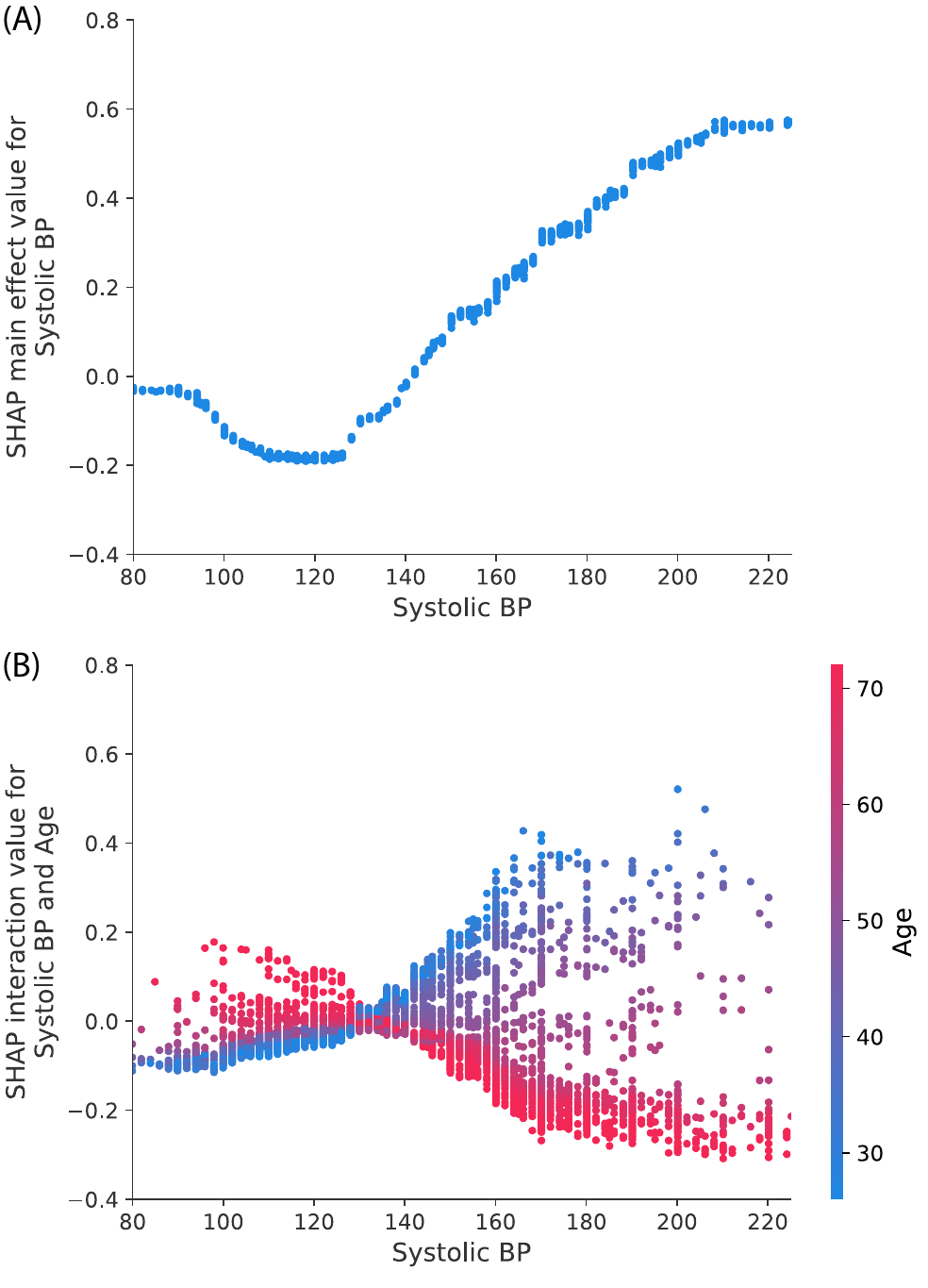}
  \caption{SHAP interaction values separate the impact of systolic blood pressure into main effects (A; Equation~\ref{eq:shapley_interaction_main}) and interaction effects (B; Equation~\ref{eq:shapley_interactions}). Systolic blood pressure has a strong interaction effect with age, so the sum of (A) and (B) nearly equals Figure \ref{fig:nhanes_sbp}. There is very little vertical dispersion in (A) since all the interaction effects have been removed.}
  \label{fig:nhanes_sbp_main_and_age_interaction}
\end{figure}

Combining SHAP dependence plots with SHAP interaction values can reveal global interaction patterns. Figure~\ref{fig:nhanes_sbp_main_and_age_interaction}A plots the SHAP main effect value for systolic blood pressure. Since SHAP main effect values represents the impact of systolic blood pressure after all interaction effects have been removed (Equation~\ref{eq:shapley_interaction_main}), there is very little vertical dispersion in Figure \ref{fig:nhanes_sbp_main_and_age_interaction}A. Figure~\ref{fig:nhanes_sbp_main_and_age_interaction}B shows the SHAP interaction value of systolic blood pressure and age. As suggested by the coloring in Figure~\ref{fig:nhanes_sbp}, this interaction accounts for most of the vertical variance in the systolic blood pressure SHAP values.

\section{Conclusion}
Several common feature attribution methods for tree ensembles are inconsistent, meaning they can lower a feature's assigned importance when the true impact of that feature actually increases. This can prevent the meaningful comparison of feature attribution values. In contrast, SHAP values consistently attribute feature importance, better align with human intuition, and better recover influential features. By presenting the first polynomial time algorithm for SHAP values in tree ensembles, we make them a practical replacement for previous methods. We further defined SHAP interaction values as a consistent way of measuring potentially hidden pairwise interaction relationships. Tree SHAP's exponential speed improvements open up new practical opportunities, such as supervised clustering, SHAP summary plots, and SHAP dependence plots, that advance our understanding of tree models.

\vspace{3.2pt}
\noindent {\it Acknowledgements:} Vadim Khotilovich for helpful feedback. 

\bibliographystyle{ACM-Reference-Format}
\bibliography{main}

\end{document}